\documentclass{article}
\usepackage{amssymb}
\usepackage{graphicx}%
\usepackage[subtle]{savetrees}

\usepackage[final]{corl_2025} 

\title{Switch4EAI: Leveraging Console Game Platform for Benchmarking Robotic Athletics}

\author{
  Tianyu Li, Jeonghwan Kim, Wontaek Kim, Donghoon Baek, Seungeun Rho, Sehoon Ha\\
  Georgia Institute of Technology, 
  USA\\
  \texttt{tli471@gatech.edu} \\
}

\begin{document}
\maketitle

\vspace{-1.cm}
\begin{abstract}
   Recent advances in whole-body robot control have enabled humanoid and legged robots to execute increasingly agile and coordinated movements. However, standardized benchmarks for evaluating robotic athletic performance in real-world settings and in direct comparison to humans remain scarce. We present Switch4EAI(Switch-for-Embodied-AI), a low-cost and easily deployable pipeline that leverages motion-sensing console games to evaluate whole-body robot control policies. Using Just Dance on the Nintendo Switch as a representative example, our system captures, reconstructs, and retargets in-game choreography for robotic execution. We validate the system on a Unitree G1 humanoid with an open-source whole-body controller, establishing a quantitative baseline for the robot's performance against a human player. In the paper, we discuss these results, which demonstrate the feasibility of using commercial games platform as physically grounded benchmarks and motivate future work to for benchmarking embodied AI.
\end{abstract}



Recent advances in whole-body robot control have enabled humanoid and legged robots to perform increasingly agile and coordinated movements. Modern control and learning techniques—ranging from model-based whole-body controllers~\citep{khazoom2024tailoring, li2024cafe, li2025gait, baek2025whole} to reinforcement learning~\citep{xie2025kungfubot, chen2025gmt, li2025amo, ze2025twist}—have demonstrated impressive capabilities, enabling robots to perform athletic activities such as gymnastic movement~\citep{}, dynamic dancing~\citep{chen2025gmt}, and real-time whole-body teleoperation~\citep{ze2025twist}. These developments mark a significant step toward robots that can interact with their environments in a more versatile and robust manner. However, as athletic capabilities continue to improve, there is a growing need for standardized evaluations that can quantify the athletic performance of these robot control algorithms.

Several recent works have leveraged large-scale human motion datasets, such as AMASS~\citep{AMASS:ICCV:2019} and LaFAN1~\citep{harvey2020robust}, to train and evaluate robot control policies with athletic capabilities. While these datasets have been instrumental in enabling diverse motion learning, the choice of training and testing splits often differs across studies, making it difficult to establish fair and reproducible comparisons between methods. Moreover, these datasets are static and infrequently updated, limiting their ability to reflect new motion patterns, environmental conditions, or emerging athletic skills. Athletic evaluations are also predominantly conducted in simulation, largely due to the complexity and cost of setting up hardware experiments, which creates a gap between simulated performance and real-world capability. Finally, existing evaluations rarely compare robot performance directly against human baselines, leaving unclear how closely robotic capabilities approach human athletic ability. These limitations highlight the need for a standardized, dynamic, and physically grounded evaluation framework that enables consistent benchmarking and direct human-robot performance comparison.

\begin{figure}
    \centering
    \includegraphics[width=0.85\linewidth]{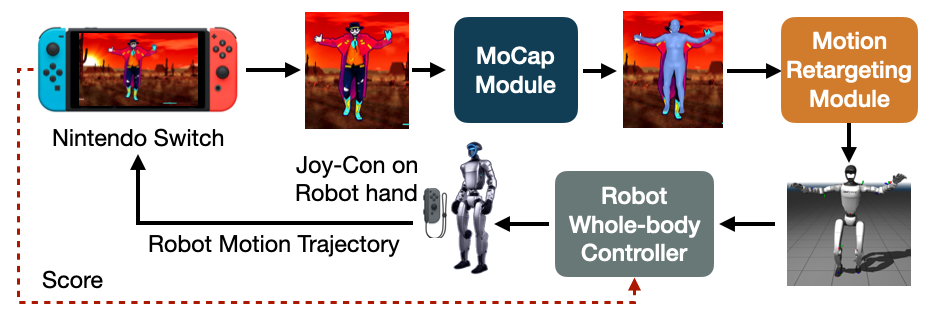}
    \caption{The Switch4EAI Pipeline. The system captures gameplay from a Nintendo Switch. The video stream is fed into the MoCap Module, which reconstructs the dancer's 3D pose. This pose is then mapped to the robot's body by the Motion Retargeting Module. The Robot Whole-body Controller executes the motion, and the robot's performance is scored in-game via Joy-Con controllers attached to its hands.}
    \label{fig:method_overview}
    \vspace{-.7cm}
\end{figure}

Commercial motion-sensing console platforms, such as the Nintendo Switch~\citep{nintendoswitch}, offer a wide range of games that require players to engage in full-body motion, with Just Dance~\citep{justdance} being a prominent example. Just Dance is a rhythm-based game where players imitate on-screen choreography, and their performance is scored in real time through motion tracking. We propose using such platforms as standardized tools for evaluating robotic athletic capabilities. These systems are inexpensive($\sim$400USD), easy to set up, and designed for repeated, long-term use. Many games receive frequent content updates, introducing new motion sequences that can serve as fresh evaluation tasks without additional data collection effort. Furthermore, these games are built to accommodate players of varying sizes, enabling fair evaluation of robots with different morphologies—from small humanoids to full-scale bipedal robots—within the same scoring framework. Finally, because these platforms are designed for human users, they naturally allow for direct performance comparison between robots and humans, providing a meaningful benchmark for assessing progress in whole-body control.
\vspace{-0.2cm}
\section*{Pipeline and Validation}
\vspace{-0.2cm}

To enable a robot to play Just Dance, we developed Switch4EAI, a pipeline designed to convert the Nintendo Switch’s visual output into whole-body control targets for humanoid robots. An overview of the pipeline is presented in Figure~\ref{fig:method_overview}. The pipeline consists of three main modules. The Streaming Module captures the Switch’s screen output in real time using an HDMI capture card and optionally processes the images for downstream motion analysis. The MoCap Module extracts human motion from the streamed video using ROMP~\citep{ROMP}, an open-source motion capture framework that reconstructs 3D SMPL~\citep{SMPL:2015} humanoid character poses from monocular RGB input. The Motion Retargeting Module then maps the SMPL motion to the robot’s morphology using GMR~\citep{ze2025gmr}, a real-time motion retargeting system. During gameplay, a pair of Joy-Con controllers is securely attached to the robot’s hands, allowing the Switch to track the robot’s movements and assign in-game scores. The robot executes the retargeted whole-body motions through its control policy, enabling direct performance evaluation within the game environment. We will open source our code after this paper gets accepted. 

We validated the Switch4EAI pipeline on a Unitree G1 humanoid robot by evaluating the open-source GMT~\citep{chen2025gmt} whole-body control policy. The robot was tested on three “easy” level (2-star) dance routines from Just Dance 2020. In each trial, the Switch4EAI system captured the reference motions from the game, retargeted them to the G1’s morphology, and executed them through the control policy while holding Joy-Con controllers to receive in-game scores. Demonstrations of the robot playing the game are provided in the supplementary video. Our results show that the pipeline successfully enabled the robot to interpret and attempt to follow the reference dance motions directly from the Switch input. We quantitatively evaluated performance by averaging the in-game scores across the three routines: the robot achieved a mean score of 5,707, compared to 9,361 for a human player. Notably, this result does not represent the upper bound of the robot’s capability, as additional engineering efforts could further improve policy performance and hardware execution.
\vspace{-0.2cm}
\section*{Conlusion and Future Work}
\vspace{-0.2cm}

In summary, we propose benchmarking robotic athletic capabilities using motion-sensing console games, with Just Dance on the Nintendo Switch serving as a representative example. Our proposed framework offers several key advantages: it is low-cost, easy to deploy, adaptable to different robot morphologies, and inherently supports direct comparison between human and robot performance. We developed an initial system enabling a Unitree G1 humanoid to participate in Just Dance and validated the feasibility of the approach by quantifying its performance using the game’s built-in scoring system. For future work, we plan to apply our benchmarking metrics to a wider range of existing whole-body controllers to enable a comprehensive performance comparison. In addition, we believe that console games have the potential to evaluate not only policy agility but also broader Embodied AI capabilities. To this end, we aim to expand to other motion-based games and explore additional opportunities offered by this platform for robotics evaluation.


\clearpage


\bibliography{reference}  

\end{document}